# Fraud Analytics Using Machine-learning & Engineering on "Big Data" (FAME) for Telecom

**Sudarson Roy Pratihar**

**Student, Indian School of Business**

**Bangalore, India**

Email: sudarson_pratihar_2014@cba.isb.edu

**Subhadip Paul**

**Student, Indian School of Business**

**Bangalore, India**

Email: subhadip_paul_2014@cba.isb.edu

**Pranab Kumar Dash**

**Student, Indian School of Business**

**Bangalore, India**

Email: pranab_dash_2014@cba.isb.edu

**Amartya Kumar Das**

**Student, Indian School of Business**

**Kolkata, India**

Email: amartya_das_2014@cba.isb.edu

*Abstract: Telecom industries lose globally 46.3 Billion USD due to fraud. Data mining and machine learning techniques (apart from rules oriented approach) have been used in past, but efficiency has been low as fraud pattern changes very rapidly. This paper presents an industrialized solution approach with self adaptive data mining technique and application of big data technologies to detect fraud and discover novel fraud patterns in accurate, efficient and cost effective manner. Solution has been successfully demonstrated to detect International Revenue Share Fraud with <5% false positive. More than 1 Terra Bytes of Call Detail Record from a reputed wholesale carrier and overseas telecom transit carrier has been used to conduct this study.*

*Keywords: Telecommunication Fraud, Fraud analytics, Big data analytics, Anomaly detection, Hadoop, Spark .*

# 1 Introduction

## 1.1 Telecom Frauds

Motive behind telecom fraud, in most cases, is to generate monetary benefits illegally for fraudsters themselves. Reference [1] discusses various types of frauds by types and methods and their impact. Table 1-1 is a snapshot from reference [1].

With advent of technologies, telecom frauds are on rise with new patterns and mechanisms. Table 1-2 below gives an example of one such fraud - *International Revenue Share Frauds* which has global impact of 10.76 Billion Dollars.

*Table 1-1 Quick Facts on Telecom Frauds*

| Global Picture | Global Impact of Top Telecom Frauds |
|---|---|
| • $46.3 Billion (USD) annual revenue loss due to fraud in telecom. | • $10.76 Billion –International Revenue Share Fraud(IRSF) |
| • 2.09% of global telecom revenue is fraud. | • $ 5.97 Billion – Interconnect Bypass (e.g. SIM Box) |
| • 375,161 fraud cases reported annually only in North America and Western Europe. | • $3.77 Billion – Premium Rate Service |
| • 816 mobile networks operators worldwide. | • $2.94 Billion – Arbitrage • $ 2.84 B – Thed / Stolen Goods |
| • 10% of operators have bad debt only due to fraud. | • $2.35 Billion – Device / Hardware Resale |
| • 46% of fraud happens over fixed and prepaid mobile services. | • $2.09 Billion – Domes-c Revenue Share (DRSF) |
| • 12% fraud happens over wholesale services | • $2.01 Billion – Wholesale Fraud |

*Table 1-2 Brief on International Revenue Sharing Fraud*

| How does it happen? | Who are impacted? |
|---|---|
| • Fraudster creates premium rate numbers in high priced destinations such as Somalia, Latvia, etc. And advertises to encourage people to call-in these numbers. Most of the time, a payment agreement is arranged to create traffic to these numbers.<br>• People with intension to gain easy money access telephony system illegally (e.g. by PBX hacking, sim cloning, stolen sim) to generate traffic to these premium rate numbers and keep the connection alive for longer duration. | • Customers – huge bill to retail and corporate customers due to fraud. Typically deny to pay.<br>• Telecom operators – International Telecom Agreement needs the operator to pay the next operator in chain of the network. So destination telecom operator is benefitted as they get their share of revenue, operators at source and transit carriers at risk of denial of payment and legal hassles. |

1.2 Current Problem

In past years, there have been several attempts at analyzing and taking preventive measures against fraudulent activities through rule bases algorithms as well as data mining. But the problem lies in the fact that the pattern of fraud changes over time with fraudsters adopting newer methods, historical pattern detection mining mechanisms are not efficient. And also volume of data (in order of a few hundred GBs in a day) makes it difficult to process data in timely manner. This paper presents an industrialized solution approach with self adaptive machine learning and application of big data technologies to address this. The approach has been demonstrated for detection of *International Revenue Sharing Fraud* (IRSF) with less than <5% false positive results. However, this approach is extendible for detection of most of telecom frauds and frauds of similar nature. More than 1.2 TB of history Call Detail Record (CDR) data from one of the top overseas transit carrier has been used to demonstrate this approach successfully.

## 2 Related Work

Reference [2] discusses user profiling based approach. However this approach cannot be extended to larger set of frauds and it will have problem when pattern changes.

Reference [3] discusses SDVAR based change point detection in time series. However as the telecom call patterns can change over time, this will require frequent recalibration.

# 3  Overall Approach

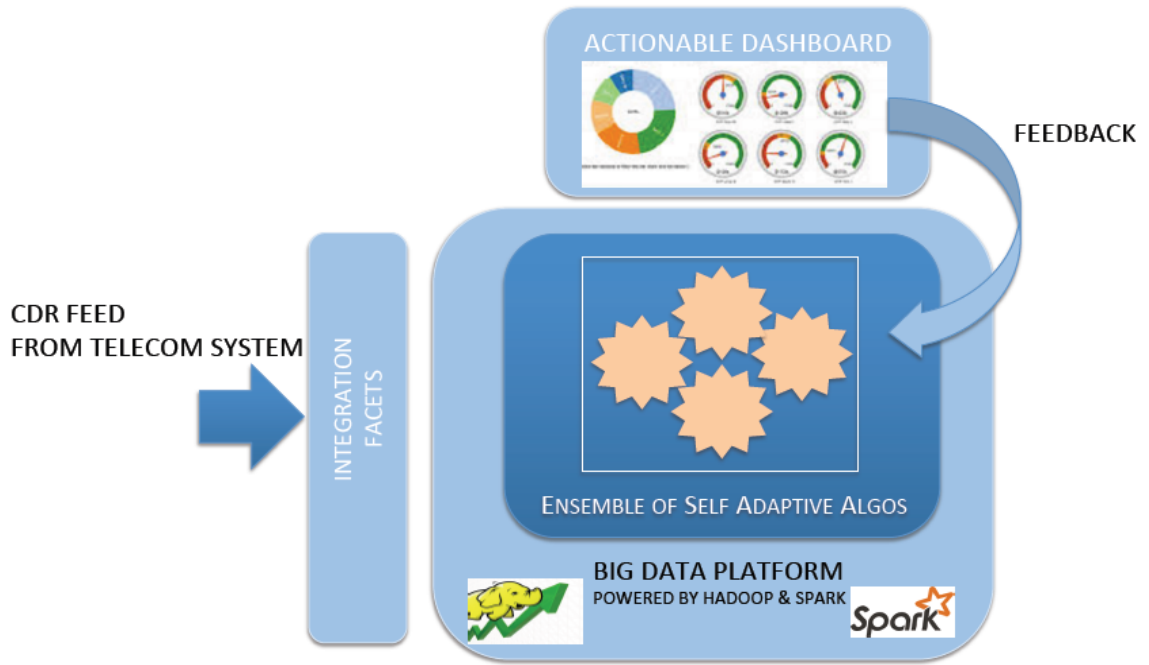

*Figure 3-1 Overview of FAME for Telecom*

Basic components of FAME are:

- Self adaptive Machine learning methodology
- Actionable dash board for operations and investigations team to act upon the alerts and feedback sent to machine learning model for adjusting weights.
- High performance big data platform for data processing and machine learning

## 3.1 Machine Learning Methodology

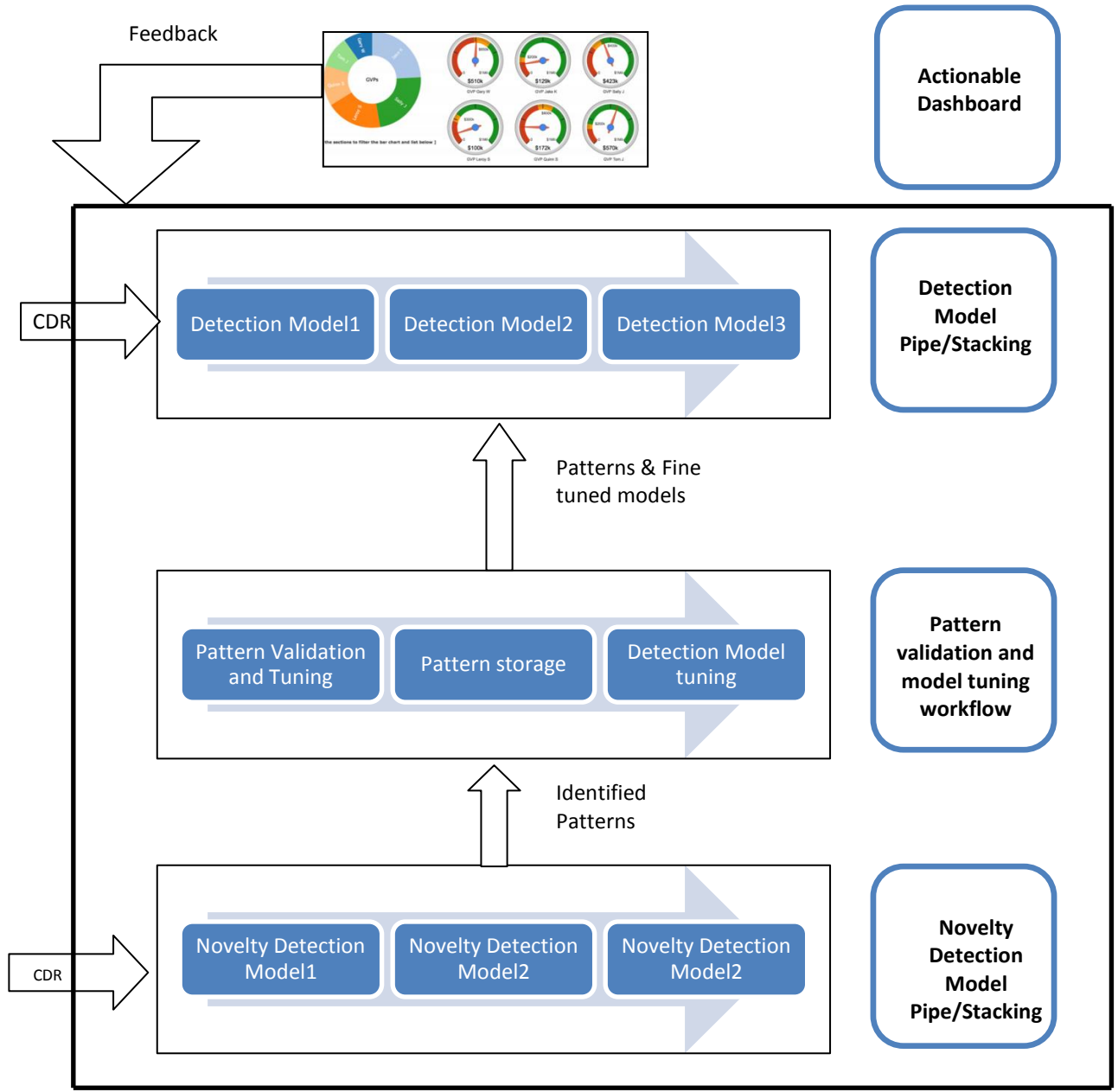

*Figure 3-2 Machine Learning Methodology*

Machine learning methodology for the approach has 3 different components:

Novelty Detection / Pattern Discovery Pipeline/Stacking

A series of novelty detection algorithms are executed to learn unknown pattern of call behavior. Pipeline combines the output of each components through a tunable logic to provide final output.

Contextual Anomaly Detection [3] is used to detect novelty in individual algorithms. In contextual anomaly, data points may be normal in other scenarios, but in presence of values of other attribute it is unusual.

Example: if a caller calling in every alternate minute to a particular destination number and total call duration is 500 min for a given time slice, this may be a contextual anomaly if majority of other callers do not do that in that point in time. However just calling for 500 min itself does not make it anomaly in absence of other context attributes.

Various algorithms have been used to detect anomaly based on type of the anomaly we want to detect, some of them are:

- Outlier based on similarity/distance matrix
- Multivariate probability distribution of normal behavior
- Empirical distribution based on historical profiling of attributes

Pattern validation and model tuning workflow

Discovered patterns are reported through an actionable dashboard and workbench. Based on feedback on the dashboard, appropriate fraud data is used to either build fresh model or tune existing models.

Tuned models are tested against the historical data and patterns before they are pushed to fraud detection model pipeline.

Patterns as well as detected fraud source or destinations are stored and used for smart filtering of future detection.

Detection Model Pipeline

Detection models are created based on fraud patterns detected. Pipeline combines the output of each components through a tunable logic to provide final output.

Models are combination of supervised simple models as well as simple threshold based models.

In this methodology, combination of small and simple algorithms has been used to achieve better performance and better accuracy.

3.2  Actionable Dashboard

For the fraud detection system to be adaptive, it is needed that feedback goes to the model on the fraud alerts generated and novel patterns discovered. System can be auto-tuned or tuned under human intervention on case-to-case basis. Actionable Dashboard facilitates this along with serving as alert and pattern reporting mechanism.

3.3  Big Data Platform

Big Data platform are leveraged to process CDR in streaming or batch mode and run machine learning algorithms to produce alerts in time efficient and cost effective manner. Combination of Apache Hadoop and Apache Spark is an example of such big data platform, however it is not restricted to only this combination.

**4  Description of Novelty Detection Pipeline**

Below are the details of novelty detection pipeline used for pattern detection in destination phone number and origin phone number for IRSF. However approach is generic and different other models could be also used for the same.

4.1  Destination Number Patterns

Fraudulent destination number (also known as B-Numbers) patterns are detected by combining output from each of below algorithms

- Calculate inter-quartile range of current data and find out distance of data points from inter-quartile range. Use a threshold parameter to predict anomaly.

- Cluster observations based on key attributes. To find out right number of clusters, calculate silhouette score. Values of centroids of the clusters will point out the cluster with new set of frauds.

4.2  Origin Number Patterns

Fraudulent origin number (also known as A-Numbers) patterns are detected by through below pipeline:

- Profile historical call data by category of source of the calls. Category could be geographical region.  In present case, it was the retail telecom operator who was passing the call to the wholesale operator. Get profiles of aggregated / rolling aggregated data of various time slots of the day of month.

- Monitoring ongoing call records against these profiles and deviations beyond threshold are put under scanner.
- Drill down these records for invidual source number analysis. Mahalonbis distance of the call attributes are used as mechanism to detect outlier.

Mahalanobis distance is a measure of the distance between a point P and a distribution D. Mahalanobis Distance is given by the formulae:

$$d(\vec{x}, \vec{y}) = \sqrt{(\vec{x}-\vec{y})^T S^{-1} (\vec{x}-\vec{y})}. \qquad (1)$$

Mahalonbis distance is appropriate measure of distance as some of the call attributes may have high degree of correlations.

Mahalonbis distances are compared against inter-quartile distances of distribution of Mahalonbis distance in following manner

$$\text{Mdist} > \phi . \text{IQR} \qquad (2)$$

$\phi$ is tuned based on historical patterns and actions through Pattern Validation WorkBench and Actionable Dash Board.

## 5  Description of Fraud Detection Pipeline

Below are the details of fraud detection pipeline used for fraud detection in destination phone number and origin phone number for IRSF. However approach is generic and different other models could be also used for the same.

By combining outputs of each algorithm, origin and destination phone numbers involved in fraudulent activities are detected:

- History – Historical fraud repository is screened to find a close match with number pattern. Typically a block of phone numbers in same series is used for fraud.
- Threshold based – Based on historical patterns threshold are defined for various attributes. Deviations are compared against predefined threshold.

- Logistic regression models – Logistic regression models developed in *Pattern Validation and Model Tuning* steps are used here. Based on pattern, appropriate models are applied.

## 6   Big Data Platform

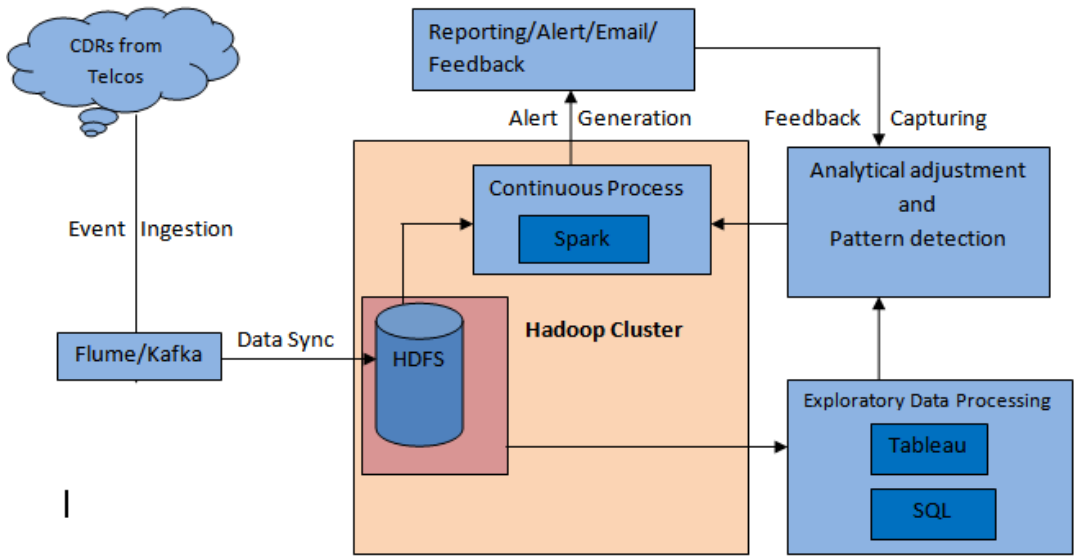

*Figure 2 High level architecture of a telecom fraud detection system*

A huge number of call data records (CDR) have to be processed and analyzed on near real time (NRT) basis for a telecom fraud detection system. The CDRs arrive as multiple small files 24X7 and swells at 10 GB of data/day. Even though the data is structured the huge proliferation of the data needs a low cost horizontally scalable environment for storing and analyzing the data.

Hadoop has become the dominant paradigm for big data processing because Hadoop Distributed File System (HDFS) is the right storage platform for big data and its capability for the distributed and parallel computing with MapReduce.  Although Hadoop provides a filesystem client that makes it easy to copy files in and out of Hadoop, most of the applications implemented on Hadoop involve ingestion of disparate data types from different sources and with differing requirements for frequency of ingestion. To meet the requirement of near real time data ingestion, Hadoop ecosystem provides tools like Flume and Kafka.

MapReduce lacks in term of 'in-memory' data processing capability and here UC Berkeley's big data processing framework Spark has an edge over MapReduce. Spark handles most of its operations "in memory" – copying them from the distributed physical storage into far faster logical RAM memory. This reduces the amount of time consuming writing and reading to and from slow, clunky mechanical hard drives that needs to be done under Hadoop's MapReduce system. To meet the NRT requirement (where processing is typically done in a few minutes after the event occurred with the intention of finding fraudulent call record) for the data cleaning, processing and analysis, Apache Spark is used for the features extraction from stored CDRs, customer level profiling and finally finding fraudulent record from the incoming CDRs to generate the fraud alerts which are given back to a reporting system for further action.

## 7 Experimental Results

Below are a few sample test results to demonstrate the theory discussed so far. Tests have been done on real telecom CDR from a telecom wholesale carrier and transit operator.

### 7.1 Origin Number Fraud and Pattern Detection

Profiling
Sample output for profiling for each telecom operator (Masked data)

| Operator | Day | Block | Mean_Y | Mean_X | SD Y | SD_X |
|---|---|---|---|---|---|---|
| XXXX | Thursday | 00:00-06:00 | 37 | 19022 | 2.12132 | 2175.768 |
| XXXX | Thursday | 01:00-07:00 | 215 | 6637 | 85.55992 | 2107.885 |
| XXXX | Thursday | 02:00-08:00 | 72 | 5894 | 9.192388 | 2107.885 |
| XXXX | Thursday | 03:00-09:00 | 184 | 4309 | 25.96194 | 2048.488 |
| XXXX | Thursday | 04:00-10:00 | 199.5 | 3798.5 | 35.35534 | 1886.561 |
| XXXX | Thursday | 05:00-11:00 | 8814.5 | 109074.5 | 462.76 | 151847.6 |

*Table 7-1 Showing mean(μ) and sd(σ) for each opeator for particular destination for specific hour blocks.*

### Monitoring & filtering

Sample output for Deviation of individual A-Numbers(masked)

| Operator Code | X | Z | Y | Deviation X | Deviation Y |
|---|---|---|---|---|---|
| 1111 | 425.2333 | 69 | 32 | 4.849776 | 5.939697 |
| 3333 | 3020.65 | 792 | 468 | 7.90914 | 13.26667 |
| 4444 | 107.7 | 27 | 14 | 2.180501 | 23.334524 |
| 5555 | 1085.183 | 65 | 54 | 4.203986 | 4.596194 |
| 6666 | 674.8 | 62 | 59 | 3.383827 | 0.989949 |
| 7777 | 83.73333 | 56 | 19 | 6.7686 | 2.592725 |
| 88888 | 102 | 4 | 4 | 6.7686 | 2.592725 |
| 101010 | 235.4833 | 33 | 25 | 6.7686 | 2.592725 |
| 7789 | 165 | 8 | 7 | 16.059304 | 7.778175 |
| 1234 | 487.1167 | 37 | 28 | 10.945766 | 0.972272 |

*Table 7-2 showing deviation from profiled data for each operator for that particular hour block.*

Sample output for Mahalanobis Distance(Mdist)

| Operator **1234** for Destination Latvia | | | | | | |
|---|---|---|---|---|---|---|
| Origin Number | X1 | X2 | X3 | X4 | X5 | m_dist |
| XXXXX | 3 | 110.3667 | 36.7889 | 1 | 2 | 3.205858 |
| XXXXX | 2 | 90.95 | 45.475 | 1 | 1 | 3.278858 |
| XXXXX | 2 | 88.11667 | 44.0583 | 1 | 2 | 3.237759 |
| XXXXX | 1 | 70.4 | 70.4 | 1 | 1 | 3.549781 |
| XXXXX | 12 | 66.6 | 8.325 | 0.66667 | 5 | 5.633712 |

*Table 7-3 showing the Mdist for each A-Number for a particular operator*

| Operator **4444** for Destination Latvia | | | | | | |
|---|---|---|---|---|---|---|
| Origin Number | X1 | X2 | X3 | X4 | X5 | **m_dist** |
| 46317498823 | 10 | 74.48333 | 8.275926 | 0.9 | 3 | 4.086288 |
| 37126110536 | 2 | 35.61667 | 35.61667 | 0.5 | 2 | 1.820313 |
| 46765422969 | 4 | 33.36667 | 8.341667 | 1 | 1 | 3.056175 |
| 4.48E+11 | 3 | 21.18333 | 7.061111 | 1 | 1 | 2.850226 |
| 31186210010 | 3 | 20.38333 | 6.794444 | 1 | 1 | 2.709994 |

*Table 7-4 showing the Mdist for each A-Number for a particular operator*

| Operator **5555** for Destination Latvia | | | | | | |
|---|---|---|---|---|---|---|
| Origin Number | X1 | X2 | X3 | X4 | X5 | **m_dist** |
| 33951428784 | 3 | 191.9333 | 63.97778 | 1 | 1 | 3.273819 |
| 33674682360 | 9 | 169.3 | 18.81111 | 1 | 5 | 4.415898 |
| 33663706311 | 2 | 121.9833 | 60.99167 | 1 | 1 | 3.35735 |
| 33782695588 | 4 | 104.5833 | 34.86111 | 0.75 | 1 | 3.337683 |
| 33666901596 | 1 | 90.68333 | 90.68333 | 1 | 1 | 3.709531 |

*Table 7-5 showing the Mdist for each A-Number for a particular operator*

**Plotting the outliers based on Mdist**

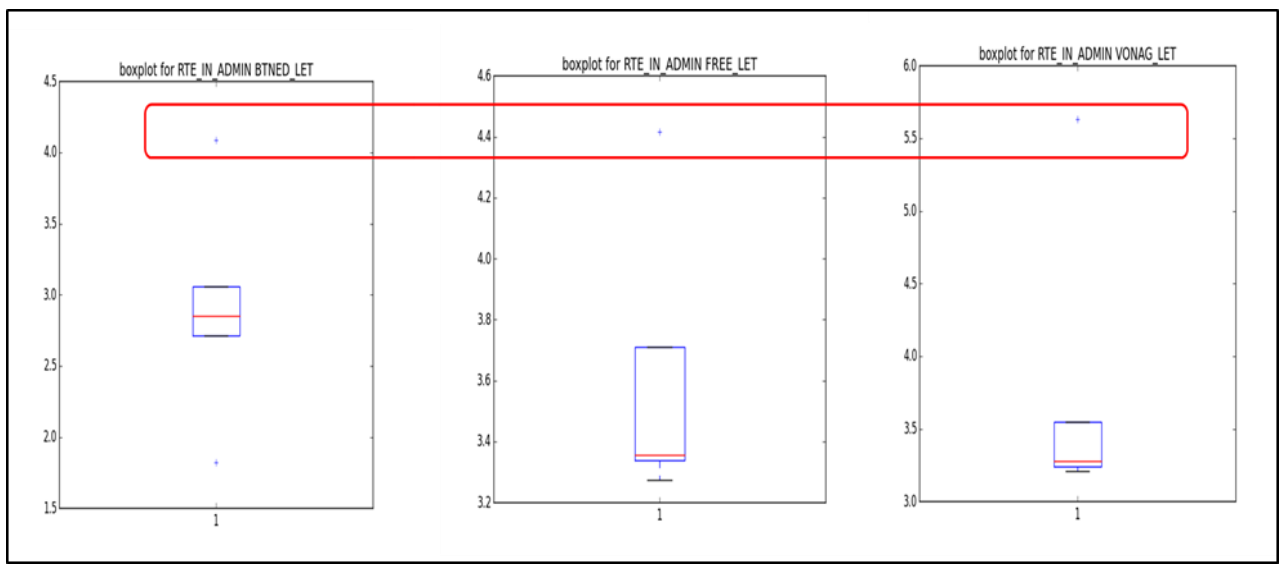

*Figure 7-1 Boxplots for Outliers based on MDist*

## 7.2 Destination Number Fraud and Pattern Detection

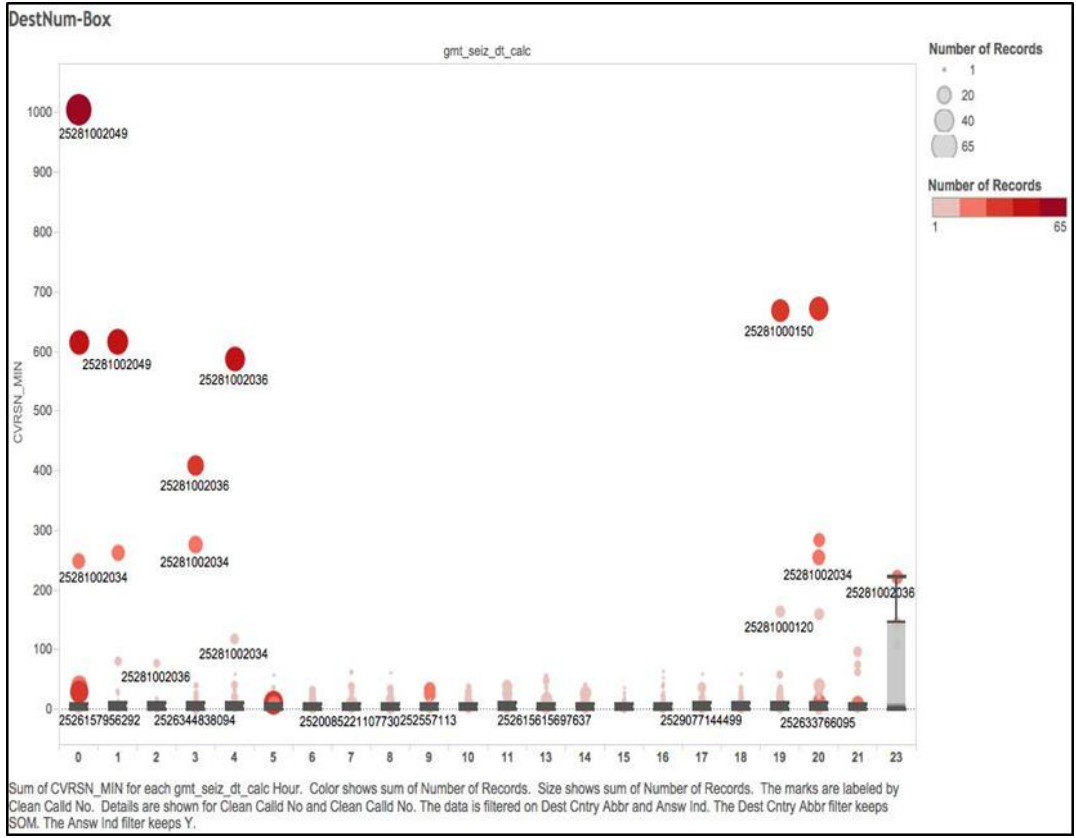

*Figure 7-2 above picture depicts hourly aggregated boxplots for each B-Numbers for a Particular destination. It is seen that there are few B-Numbers which are plotted as outliers as compared to others numbers.*

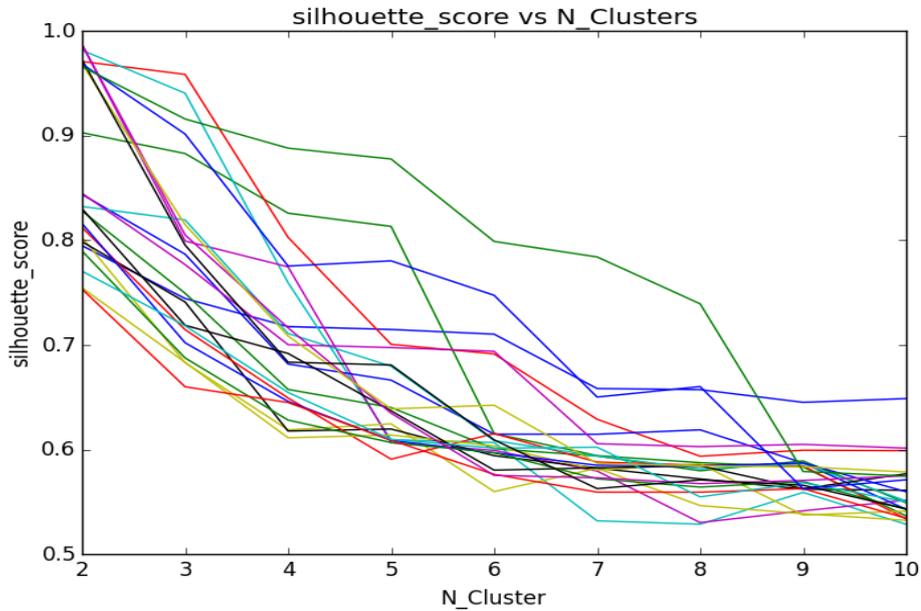

*Figure 7-3-Above image depicts the silhouette score versus number of clusters for the same hourly aggregated data. The best score is achieved when k=2.The best cluster situation has minimum silhouette score of 0.75.*

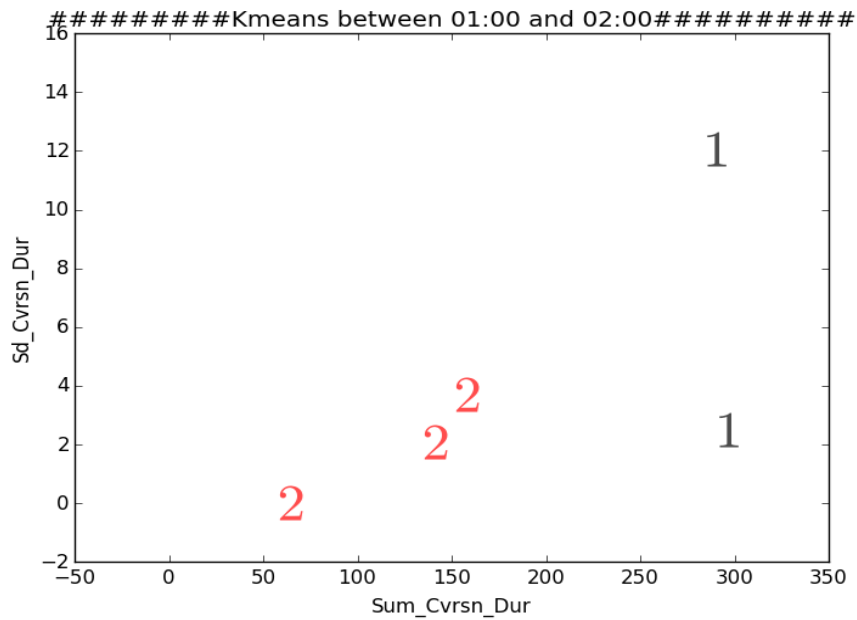

*Figure 7-4 showing 2 cluster scenario for particular hour block with two distinct clusters*

## 7.3 Alerts Generated and Hit ratio

The models are simple and fast in execution. As false positives have high cost of investigation by human, entire methodology was tuned and optimized to provide optimum false positives. Below graph shows individual rate of false positives of origin and destination number detection. However when combined, combined mechanism provided <5% false positive.

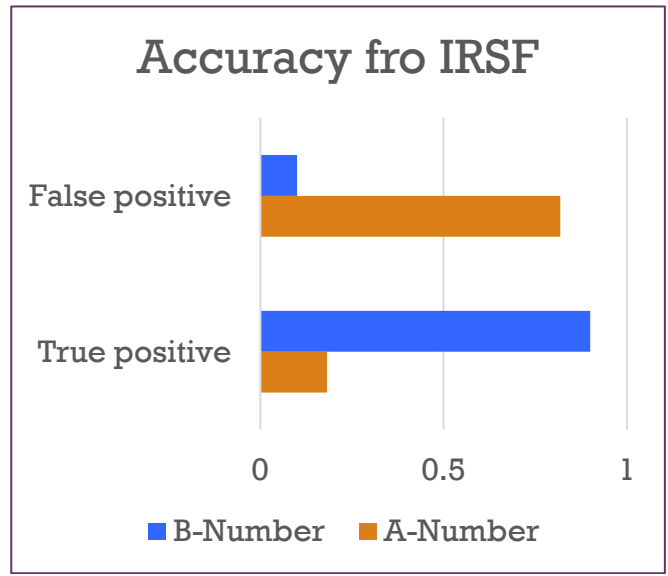

*Figure 7-5True Positive versus False Positive for A(Origin)-number and B(Destination)-number*

## 8 Conclusions
In this paper, self adaptive telecom fraud detection mechanism has been proposed along with application of big data technologies. Authors have conceived and developed this mechanism as part of their research project at Indian School of Business. Algorithms have been developed based on the research and detailed study of history call detail records by a telecom wholesale operator. Authors have also architected and developed technology platform and implemented the algorithms successfully. Results were physically investigated by the telecom wholesale operator's investigation team on live call detail records. Results were also compared against existing mechanism and software deployed for fraud detection and it was found that this methodology and algorithms are superior in performance and accuracy.